\pdfoutput=1

\documentclass[11pt]{article}

\usepackage{EMNLP2023}

\usepackage{times}
\usepackage{latexsym}
\usepackage{xcolor}

\usepackage[T1]{fontenc}

\usepackage[utf8]{inputenc}

\usepackage{microtype}

\usepackage{inconsolata}

\usepackage{graphicx}
\usepackage{amsmath}
\usepackage{amsfonts} 
\usepackage{booktabs}
\usepackage{lipsum}
\usepackage{multicol}
\usepackage{multirow}
\usepackage{bm}
\usepackage{cleveref}
\usepackage{easyeqn}
\usepackage{listings}
\usepackage{placeins}
\usepackage{color,xcolor,colortbl}
\usepackage[multiple]{footmisc}
\usepackage{enumitem}
\usepackage{paralist}

\newcommand\YAMLcolonstyle{\color{black}\mdseries}
\newcommand\YAMLkeystyle{\ttfamily\scriptsize\color{teal}\bfseries}
\newcommand\YAMLvaluestyle{\color{black}\mdseries}

\makeatletter

\newcommand\language@yaml{yaml}

\expandafter\expandafter\expandafter\lstdefinelanguage
\expandafter{\language@yaml}
{
  keywords={},
  keywordstyle=\color{black}\mdseries,
  basicstyle=\YAMLkeystyle,
  sensitive=false,
  comment=[l]{\#},
  morecomment=[s]{/*}{*/},
  commentstyle=\color{purple}\ttfamily,
  stringstyle=\YAMLvaluestyle\ttfamily,
  moredelim=[l][\color{orange}]{\&},
  moredelim=[l][\color{magenta}]{*},
  moredelim=**[il][\YAMLcolonstyle{:}\YAMLvaluestyle]{:},   
  morestring=[b]',
  morestring=[b]",
  literate =    {---}{{\ProcessThreeDashes}}3
                {>}{{\textcolor{red}\textgreater}}1     
                {|}{{\textcolor{red}\textbar}}1 
                {\ -\ }{{\mdseries\ -\ }}3,
  belowskip=-3.0 \baselineskip
}

\lst@AddToHook{EveryLine}{\ifx\lst@language\language@yaml\YAMLkeystyle\fi}
\makeatother

\newcommand\ProcessThreeDashes{\llap{\color{cyan}\mdseries-{-}-}}

\newcommand{\uv}{\mathbf{u}}
\newcommand{\vv}{\mathbf{v}}
\newcommand{\xv}{\mathbf{x}}
\newcommand{\yv}{\mathbf{y}}
\newcommand{\thetav}{\bm{\theta}}
\newcommand{\EE}{\mathbb{E}}

\newcommand{\DC}{\mathcal{D}}
\newcommand{\CC}{\mathcal{C}}
\newcommand{\HC}{\mathcal{H}}
\newcommand{\IC}{\mathcal{I}}
\newcommand{\KC}{\mathcal{K}}
\newcommand{\MC}{\mathcal{M}}
\newcommand{\YC}{\mathcal{Y}}

\newcommand{\multirowcell}[1]{\begin{tabular}[c]{@{}c@{}}#1\end{tabular}}

\newcommand\blankfootnote[1]{%
  \let\thefootnote\relax\footnotetext{#1}%
  \let\thefootnote\svthefootnote%
}

\title{LM-Polygraph: Uncertainty Estimation for Language Models}

\author{
\bf Ekaterina Fadeeva\textsuperscript{3,5 $\diamondsuit$}\enspace
Roman Vashurin\textsuperscript{2 $\diamondsuit$}\enspace Akim Tsvigun\textsuperscript{5,6,7 $\diamondsuit$}\enspace Artem Vazhentsev\textsuperscript{3,4 $\diamondsuit$} \\
\bf Sergey Petrakov\textsuperscript{3 $\diamondsuit$}\quad 
Kirill Fedyanin\textsuperscript{2}\quad 
Daniil Vasilev\textsuperscript{5}\quad Elizaveta Goncharova\textsuperscript{4,5,6} \\ 
\bf Alexander Panchenko\textsuperscript{3,4}\quad
 Maxim Panov\textsuperscript{1}\quad
Timothy Baldwin\textsuperscript{1,8}\quad
Artem Shelmanov\textsuperscript{1}\\
\textsuperscript{1}MBZUAI\quad
\textsuperscript{2}TII\quad
\textsuperscript{3}Skoltech\quad 
\textsuperscript{4}AIRI\quad
\textsuperscript{5}HSE University
\\
\textsuperscript{6}AI Center NUST MISiS\quad
\textsuperscript{7}Semrush\quad
\textsuperscript{8}The University of Melbourne
\\
\href{mailto:ekaterina.fadeeva@skol.tech}{\{ekaterina.fadeeva, sergey.petrakov\}@skol.tech} ~~
\href{mailto:roman.vashurin@tii.ae}{\{roman.vashurin, kirill.fedyanin\}@tii.ae}\\ 
\href{akim.tsvigun@semrush.com}{akim.tsvigun@semrush.com} ~~
\href{mailto:vazhentsev@airi.net}{\{vazhentsev, panchenko, goncharova\}@airi.net} \\
\href{mailto:artem.shelmanov@mbzuai.ac.ae}{\{maxim.panov, timothy.baldwin, artem.shelmanov\}@mbzuai.ac.ae}
}

\begin{document}

\maketitle

\begin{abstract}
  Recent advancements in the capabilities of large language models (LLMs) have paved the way for a myriad of groundbreaking applications in various fields. However, a significant challenge arises as these models often ``hallucinate'', i.e., fabricate facts without providing users an apparent means to discern the veracity of their statements. Uncertainty estimation (UE) methods are one path to safer, more responsible, and more effective use of LLMs. However, to date, research on UE methods for LLMs has been focused primarily on theoretical rather than engineering contributions. In this work, we tackle this issue by introducing LM-Polygraph, a framework with implementations of a battery of state-of-the-art UE methods for LLMs in text generation tasks, with unified program interfaces in Python.\footnote{\url{http://lm-polygraph.nlpresearch.group}} Additionally, it introduces an extendable benchmark for consistent evaluation of UE techniques by researchers, and a demo web application that enriches the standard chat dialog with confidence scores, empowering end-users to discern unreliable responses.\footnote{\url{http://lm-polygraph-demo.nlpresearch.group}}\footnote{\url{http://lm-polygraph-video.nlpresearch.group}} LM-Polygraph is compatible with the most recent LLMs, including BLOOMz, LLaMA-2, ChatGPT, and GPT-4, and is designed to support future releases of similarly-styled LMs. 
\end{abstract}

\section{Introduction}

\blankfootnote{$\diamondsuit$ Equal contribution}

  Large language models (LLMs) have demonstrated remarkable performance across a variety of text generation tasks. Instruction fine-tuning and reinforcement learning from human feedback (RLHF) have brought the zero-shot performance of these models to a new level~\cite{ouyang2022training}. However, the capabilities of LLMs, despite their profound power and complexity, are inherently constrained. Limitations arise from the finite nature of the training data and the model's intrinsic memorization and reasoning capacities. Hence, their utility is bounded by the depth and breadth of the knowledge they embed.

  Due to their training objectives, even when the embedded knowledge of an LLM on a given topic is limited, it tends to be over-eager to respond to a prompt, sometimes generating misleading or entirely erroneous output. This dangerous behavior of attempting to appease the user with plausible-sounding but potentially false information is known as ``hallucination''~\cite{xiao-wang-2021-hallucination,dziri-etal-2022-origin}. It poses a significant challenge when deploying LLMs in practical applications.

  There are several well-known approaches to censoring LLM outputs, including: filtering with stop-word lists, post-processing with classifiers~\cite{xu-etal-2023-understanding}, rewriting of toxic outputs~\cite{logacheva-etal-2022-paradetox}, and longer fine-tuning with RLHF. However, these approaches cannot be relied on to completely resolve hallucinations.
  Since LMs are natural (if ``unintentional'') liars, we propose LM-Polygraph --- a program framework that, similar to a human polygraph, leverages various hidden signals to reveal when one should not trust the subject. In particular, LM-Polygraph provides a comprehensive collection of uncertainty estimation (UE) techniques for LLMs in text generation tasks.

  Uncertainty estimation refers to the process of quantifying the degree of confidence in the predictions made by a machine learning model. For classification and regression tasks, there is a well-developed battery of methods~\cite{Gal2016Uncertainty}. There has also been a surge of work investigating UE, particularly in text classification and regression in conjunction with encoder-only LMs such as BERT~\cite{zhang-etal-2019-mitigating,he2020towards,shelmanov-etal-2021-certain,xin-etal-2021-art,vazhentsev-etal-2022-uncertainty,kotelevskii2022nonparametric,wang2022uncertainty}. However, UE for sequence generation tasks, including text generation, is a much more complex problem. To quantify the uncertainty of the whole sequence, we have to aggregate uncertainties of many individual token predictions and deal with non-trivial sampling and pruning techniques like beam search. Contrary to classification tasks where the number of possible prediction options is finite, in text generation, the number of possible predictions is infinite or exponential in vocabulary size, complicating the estimation of probabilities and information-based scores. Finally, a natural language text is not a simple sum of its tokens; it is a nuanced interposition of context, semantics, and grammar, so two texts can have very diverse surface forms but similar meanings, which should be taken into account during the UE process. 

  Several recent studies have delved into developing UE methods for LMs in text generation tasks~\cite{malinin2020uncertainty,van-der-poel-etal-2022-mutual,kuhn2023semantic,ren2023outofdistribution,vazhentsev-etal-2023-efficient,lin2023generating}. However, the current landscape of this research is quite fragmented with many non-comparable or even concurrent studies, which makes it challenging to consolidate the findings and draw holistic conclusions. 

  In this work, with the development of LM-Polygraph, we strive to bridge these disparate research efforts, fostering more cohesion and synergy in the field. We envision a framework that consolidates the scattered UE techniques within unified frameworks in Python, provides an extendable evaluation benchmark, and offers tools to integrate uncertainty quantification in standard LLM pipelines seamlessly.
  This endeavor will not only make the journey less challenging for individual researchers and developers but also set the stage for more robust, reliable, and trustworthy LLM deployments for end-users. 




  Our \textbf{contributions} are as follows:
  \begin{compactitem}
    \item We provide a comprehensive framework that implement state-of-the-art methods for UE of LM predictions. We also provide the ability to combine multiple uncertainty scores together as suggested by~\citet{ren2023outofdistribution,vazhentsev-etal-2023-hybrid}.

    \item We create a tool that enriches standard LLM chat capabilities with uncertainty scores for model outputs. The tool can potentially be used by end-users to determine whether the answers of language models are reliable or not, and by researchers to develop novel UE techniques for LMs in text generation tasks.

    \item We construct an easy-to-extend benchmark for UE methods in text generation tasks and provide reference experimental results for implemented UE techniques.
  \end{compactitem}

  \begin{figure}
\begin{lstlisting}[language=Python]
from lm_polygraph import estimate_uncertainty
from lm_polygraph import WhiteboxModel
from lm_polygraph.estimators import *

model = WhiteboxModel.from_pretrained(
    "bigscience/bloomz-3b",
    device="cuda:0",
)
ue_method = MeanPointwiseMutualInformation()

input_text = "Who is George Bush?"
estimate_uncertainty(model, ue_method, input_text=input_text)

# Output:
# UncertaintyOutput(
#   generation='President of the United States',
#   uncertainty=-6.858096446298684)
\end{lstlisting}
\caption{Code example of how LM predictions could be enriched with uncertainty scores using LM-Polygraph.}\label{code:polygraph}
\end{figure}

\section{Python Library}
  LM-Polygraph implements a set of state-of-the-art UE techniques for LLMs with unified program interfaces in Python. It is compatible with models from the Huggingface library and is tested with recent public-domain LLMs such as BLOOMz~\cite{scao2022bloom,yong2022bloom}, Dolly v2~\cite{DatabricksBlog2023DollyV2}, Alpaca~\cite{alpaca}, LLaMA-2~\cite{touvron2023llama}, and Flan-T5~\cite{chung2022scaling}. The framework supports both conditional models with a seq2seq architecture and unconditional decoder-only LMs. \Cref{code:polygraph} contains a code example of LM-Polygraph with BLOOMz-3B for UE in open-domain question answering. Some methods that do not require access to the model itself or its logits could be used in conjunction with web-hosted LLMs like ChatGPT or GPT-4 through APIs. We provide a program wrapper for integration with popular online services.

  \begin{table*}[t]
\centering
\small

\scalebox{0.86}{
\begin{tabular}{l|c|c|c|c|c}

\toprule
\textbf{Uncertainty Estimation Method} & \textbf{Type} & \textbf{Category} &\textbf{Compute} & \textbf{Memory} & \multirowcell{\textbf{Need} \\ \textbf{Training} \\ \textbf{Data?}} \\ 
\midrule
\midrule

Maximum sequence probability & \multirow{7}{*}{White-box} & \multirow{7}{*}{\multirowcell{Information-\\based}} & Low & Low & No \\
Perplexity \cite{fomicheva-etal-2020-unsupervised} &  & & Low & Low & No \\
Mean token entropy \cite{fomicheva-etal-2020-unsupervised} &  & & Low & Low & No \\ 
Monte Carlo sequence entropy \cite{kuhn2023semantic} &  & & High & Low & No \\
Pointwise mutual information (PMI) \cite{takayama-arase-2019-relevant} &  & & Medium & Low & No \\
Conditional PMI \cite{van-der-poel-etal-2022-mutual} &  & & Medium & Medium & No \\
\hline

Semantic entropy \cite{kuhn2023semantic} & White-box & \multirowcell{Meaning \\diversity} & High & Low & No \\
\hline

Sentence-level ensemble-based measures \cite{malinin2020uncertainty} & \multirow{3}{*}{White-box} & \multirow{3}{*}{Ensembling} & High & High & Yes \\
Token-level ensemble-based measures \cite{malinin2020uncertainty} & & & High & High & Yes \\
\hline

Mahalanobis distance (MD) \cite{lee2018simple} & \multirow{4}{*}{White-box} & \multirow{4}{*}{\multirowcell{Density-\\based}} & Low & Low & Yes \\
Robust density estimation (RDE) \cite{yoo-etal-2022-detection} &  & & Low & Low & Yes \\
Relative Mahalanobis distance (RMD) \cite{ren2023outofdistribution} &  & & Low & Low & Yes \\
Hybrid Uncertainty Quantification (HUQ) \cite{vazhentsev-etal-2023-hybrid} &  & & Low & Low & Yes \\
\hline

p(True) \cite{kadavath2022language} & White-box & Reflexive & Medium & Low & No \\
\hline

Number of semantic sets (NumSets) \cite{lin2023generating} & \multirow{5}{*}{Black-box} & \multirow{5}{*}{\multirowcell{Meaning \\diversity}} & High & Low & No \\
Sum of eigenvalues of the graph Laplacian (EigV) \cite{lin2023generating} &  & & High & Low & No \\
Degree matrix (Deg) \cite{lin2023generating} &  & & High & Low & No \\
Eccentricity (Ecc) \cite{lin2023generating} &  & & High & Low & No \\
Lexical similarity (LexSim) \cite{fomicheva-etal-2020-unsupervised} & & & High & Low & No \\

\bottomrule

\end{tabular}

}

\caption{UE methods implemented in LM-Polygraph.}
\label{tab:ue_methods}

\end{table*}

\section{Uncertainty Estimation Methods} 
Here, we summarize UE methods implemented in LM-Polygraph, as listed in \Cref{tab:ue_methods}.

There are two major technique types: white-box and black-box. The \textit{white-box} methods require access to logits, internal layer outputs, or the LM itself. The \textit{black-box} methods require access only to the generated texts, and can easily be integrated with third-party online services like OpenAI LM API. We note that the methods differ by computational requirements: some techniques pose high computational or memory overheads, e.g., due to repeated inference, making them less suitable for practical usage. The application of some methods also can be hindered by the need for access to the model training data.  
  
  Let us consider the input sequence $\xv$ and the output sequence $\yv \in \YC$ of length $L$, where $\YC$ is a set of all possible output sequences. Then the probability of an output sequence given an input sequence for probabilistic autoregressive language models is given by:
  \begin{equation}
    P(\yv \mid \xv, \thetav) = \prod\nolimits_{l = 1}^L P(y_l \mid \yv_{<l}, \xv, \thetav),
  \end{equation}
  where the distribution of each $y_l$ is conditioned on all the previous tokens in a sequence $\yv_{<l} = \{y_1, \dots, y_{l - 1}\}$, and $\thetav$ denotes the parameters of the model.

  \begin{figure*}[t]
    \centering
    \includegraphics[trim={0.cm 0.cm 0.cm 0.cm},clip,width=0.99\linewidth]{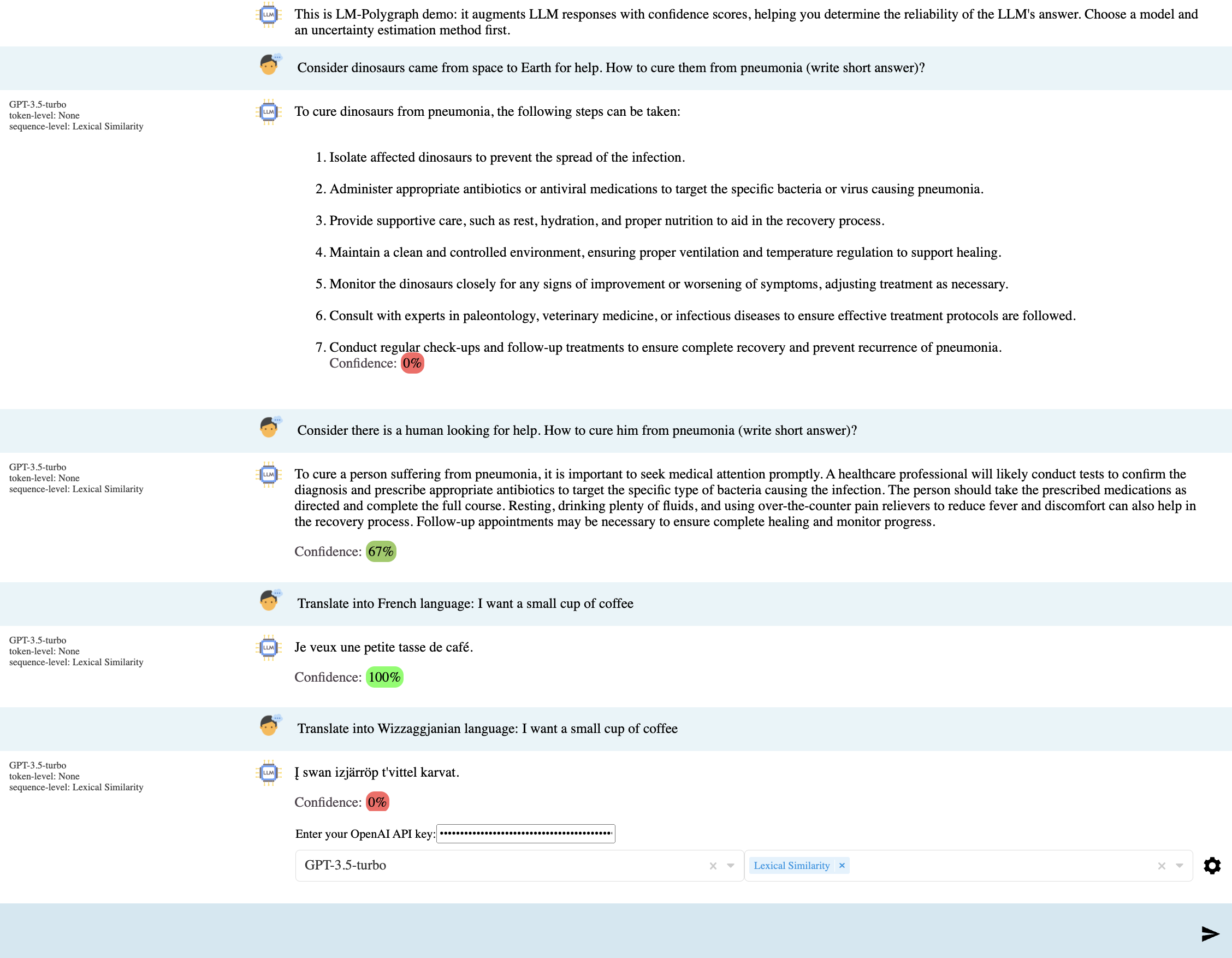}
    \caption{User interface of the demo. A user can interact with an LLM as with any other chat service, but in LM-Polygraph the user also sees the confidence of the model answers. It is possible to specify various UE techniques and various models, including ChatGPT.
    }
    \label{fig:gui}
  \end{figure*}

\subsection{White-box Methods}
  We start the discussion of white-box techniques from \textbf{information-based methods}. These techniques are based on token $P(y_l \mid \yv_{<l}, \xv, \thetav)$ and sequence $P(\yv \mid \xv, \thetav)$ probabilities obtained from a single model prediction. The notable example is \textit{entropy}, which can be calculated on the token or sequence level. The benefits of information-based methods are that they are cheap to compute and simple to implement. However, the quality of these methods is usually relatively low, so they are considered as baselines. 
  Some domain-specific methods were recently proposed in an attempt to improve over standard information-based approaches, such as \textit{semantic entropy}~\cite{kuhn2023semantic}.

  The second category of white-box techniques is \textbf{ensemble-based methods}, which leverage the diversity of output predictions made by multiple slightly different versions of models under slightly different conditions. 
  Let us assume that $M$ models are available with parameters $\thetav_i, i = 1, \dots, M$. These parameters can be obtained via independent training of models. Then one can use token $P(y_l \mid \yv_{<l}, \xv, \thetav_i)$ and sequence $P(\yv \mid \xv, \thetav_i)$ probabilities to compute various metrics such as \textit{mutual information} that measures the discrepancy between model predictions.

  \textbf{Density-based methods} leverage latent representations of instances and construct a probability density on top of them. Usually, these methods approximate training data distribution with the help of  one or multiple Gaussian distributions. They can provide a probability or an unnormalized score that determines how likely instances belong to the training data distribution. Therefore, they are good at spotting out-of-distribution (OOD) instances~\cite{vazhentsev-etal-2023-efficient}. Several variations of these methods have been proposed in the literature~\cite{lee2018simple,yoo-etal-2022-detection,ren2023outofdistribution,kotelevskii2022nonparametric}.

  The primary advantage of these methods is that they are computationally efficient: they do not need much time for additional model inference, and memory overhead for storing additional parameters is minimal. The drawback is that these methods require access to the model's training data to fit auxiliary models like Gaussians (e.g., the Mahalanobis Distance method requires constructing data centroids and covariance matrices). These methods are also known to capture only epistemic uncertainty. Therefore, they might not be perfect for selective generation as they cannot be used to spot ambiguous in-domain instances.

  Finally, we also combine information-based and density-based methods as suggested by~\citet{vazhentsev-etal-2023-hybrid} and~\citet{ren2023outofdistribution}. More specifically, we implement the \textit{hybrid uncertainty quantification (HUQ)} method~\cite{vazhentsev-etal-2023-hybrid} that performs a ranking-based aggregation and leverages strengths of both information-based methods that detect ambiguous instances and density-based methods that detect OOD instances.

  Directly \textbf{asking the model to validate its answer} is another option for UE~\cite{kadavath2022language}. In this method, one asks models first to propose answers and then to evaluate the probability $P(\text{True})$ that their answers are correct. \citet{kadavath2022language} show that it achieves reasonable performance on a variety of tasks, including question-answering. We note that this method requires inference of a model twice: the first to generate an answer, and the second for processing its own output. Even though the second inference is usually faster than the first one, it still takes considerable time for computation.

\subsection{Black-box Methods}
  In contemporary models, there are instances where the model's architecture and hidden states are unavailable or there is no access to logits during response generation. Nevertheless, a whole class of black box methods only needs to access the model's response. Within the scope of this paper, we consider several approaches of this type that have performed well in other studies~\cite{fomicheva-etal-2020-unsupervised,kuhn2023semantic,lin2023generating}. We focus on Lexical Similarity, Number of Semantic Sets, Sum of Eigenvalues of the Graph Laplacian, Degree Matrix, and Eccentricity. We use the same methodological approach as the authors of the work~\cite{lin2023generating}:
  \begin{compactitem}
    \item Obtain $K$ responses $\yv_{1}, \dots, \yv_{K}$ for a particular input $\xv$.
    \item Compute $K \times K$ similarity matrix $S$ between responses, where $S_{ij} = s(\yv_{i}, \yv_{j})$ for some similarity score $s$ (Natural Language Inference score or Jaccard score).
    \item Based on the similarity matrix $S$, we compute the final uncertainty score.
  \end{compactitem}
  Thus, the idea of the methods is to analyze the similarity matrix and aggregate the information to compute the uncertainty score.

\begin{table*}[!t] \resizebox{\textwidth}{!}{
\begin{tabular}{l|cc|cc|cc|cc|cc|cc}
\toprule
\multirow{2}{*}{\textbf{UE Method}} & \multicolumn{2}{c|}{\textbf{AESLC}} & \multicolumn{2}{c|}{\textbf{XSUM}} & \multicolumn{2}{c|}{\textbf{CoQA}} & \multicolumn{2}{c|}{\textbf{bAbiQA}} & \multicolumn{2}{c|}{\textbf{WMT14 De-En}} & \multicolumn{2}{c}{\textbf{WMT14 Fr-En}} \\ \cline{2-13}
    & \textbf{ROUGE-L} & \textbf{BERTScore}& \textbf{ROUGE-L} & \textbf{BERTScore}& \textbf{ROUGE-L} & \textbf{BERTScore}& \textbf{ROUGE-L} & \textbf{BERTScore}&  \textbf{ROUGE-L} & \textbf{BERTScore}& \textbf{ROUGE-L} & \textbf{BERTScore} \\\midrule
\midrule
            Maximum Sequence Probability &                 \cellcolor[rgb]{0.5,0.6938638985005767,0.5780392156862745} 0.24$\pm$0.01 &                 \cellcolor[rgb]{0.5,0.6333333333333333,0.5529411764705883} 0.19$\pm$0.10 &  \cellcolor[rgb]{0.8971164936562861,0.9596309111880046,0.8838523644752019} 0.01$\pm$0.03 & \cellcolor[rgb]{0.9843137254901961,0.9941176470588236,0.9803921568627452} -0.18$\pm$0.15 &                 \cellcolor[rgb]{0.5,0.6333333333333333,0.5529411764705883} 0.35$\pm$0.01 &                 \cellcolor[rgb]{0.5,0.6333333333333333,0.5529411764705883} 0.29$\pm$0.01 &                 \cellcolor[rgb]{0.5,0.6333333333333333,0.5529411764705883} 0.66$\pm$0.02 &                 \cellcolor[rgb]{0.5,0.6333333333333333,0.5529411764705883} 0.82$\pm$0.03 &  \cellcolor[rgb]{0.5540945790080738,0.7600922722029988,0.6249134948096886} 0.32$\pm$0.01 &  \cellcolor[rgb]{0.5261053440984237,0.7361014994232987,0.6049211841599385} 0.39$\pm$0.01 &  \cellcolor[rgb]{0.5648596693579393,0.7693194925028835,0.6326028450595924} 0.25$\pm$0.01 &     \cellcolor[rgb]{0.5304113802383699,0.7397923875432526,0.6079969242599} 0.31$\pm$0.01 \\
\bottomrule
\end{tabular}
}\caption{\label{tab:vicuna_results} PRR$\uparrow$ for the Vicuna model with ROUGE-L and BERTScore as text quality metrics. Darker color indicates better results.}\end{table*}
\begin{table*}[!ht] \resizebox{\textwidth}{!}{\begin{tabular}{l|cc|cc|cc|cc|cc|cc}
\toprule
\multirow{2}{*}{\textbf{UE Method}} & \multicolumn{2}{c|}{\textbf{AESLC}} & \multicolumn{2}{c|}{\textbf{XSUM}} & \multicolumn{2}{c|}{\textbf{CoQA}} & \multicolumn{2}{c|}{\textbf{bAbiQA}} & \multicolumn{2}{c|}{\textbf{WMT14 De-En}} & \multicolumn{2}{c}{\textbf{WMT14 Fr-En}} \\ \cline{2-13}
    & \textbf{ROUGE-L} & \textbf{BERTScore}& \textbf{ROUGE-L} & \textbf{BERTScore}& \textbf{ROUGE-L} & \textbf{BERTScore}& \textbf{ROUGE-L} & \textbf{BERTScore}&  \textbf{ROUGE-L} & \textbf{BERTScore}& \textbf{ROUGE-L} & \textbf{BERTScore} \\\midrule
\midrule
            Maximum Sequence Probability &                 \cellcolor[rgb]{0.5,0.6333333333333333,0.5529411764705883} 0.22$\pm$0.02 &                 \cellcolor[rgb]{0.5,0.6333333333333333,0.5529411764705883} 0.22$\pm$0.10 &                   \cellcolor[rgb]{0.5,0.711518646674356,0.585359477124183} 0.12$\pm$0.03 &   \cellcolor[rgb]{0.508881199538639,0.7213379469434833,0.5926182237600923} 0.16$\pm$0.03 &                 \cellcolor[rgb]{0.5,0.6333333333333333,0.5529411764705883} 0.44$\pm$0.01 &                 \cellcolor[rgb]{0.5,0.6333333333333333,0.5529411764705883} 0.45$\pm$0.01 &                 \cellcolor[rgb]{0.5,0.6333333333333333,0.5529411764705883} 0.43$\pm$0.03 &                 \cellcolor[rgb]{0.5,0.6459438677431757,0.5581699346405229} 0.93$\pm$0.00 &                 \cellcolor[rgb]{0.5,0.6333333333333333,0.5529411764705883} 0.44$\pm$0.01 &                 \cellcolor[rgb]{0.5,0.6333333333333333,0.5529411764705883} 0.64$\pm$0.01 &                 \cellcolor[rgb]{0.5,0.6333333333333333,0.5529411764705883} 0.45$\pm$0.01 &                 \cellcolor[rgb]{0.5,0.6333333333333333,0.5529411764705883} 0.60$\pm$0.02 \\
\bottomrule
\end{tabular}
}\caption{\label{tab:llama_2_results} PRR$\uparrow$ for the LLaMA-2 model with ROUGE-L and BERTScore as text quality metrics. Darker color indicates better results.}\end{table*}

\section{Demo}
  We constructed a demo application that can be used to interact with LLMs and also see confidence scores of model answers (see Figure~\ref{fig:gui}). 
  A user specifies a UE method and a language model from a number of publicly-available LLMs with up to 13B parameters, e.g., BLOOMz, Vicuna, and LLaMA-2. There is also the ability to communicate with LLMs deployed as web services such as ChatGPT or GPT-4 and obtain their uncertainty scores based on black-box techniques. For these models, a user should provide an API key.

  This demo application is potentially helpful for  both end-users and researchers. For end-users, it extends the standard AI assistant interface with information about whether it is reasonable to trust a model answer. Researchers could use this tool for qualitative analysis of various UE methods and LLM responses. 

\section{Evaluation Benchmark}
\label{sec:benchmark}
  LM-Polygraph provides a vast evaluation benchmark. It contains a script for running one or multiple experiments with UE techniques, implemented as Python modules. This feature allows the user to easily extend the set of available methods and evaluate novel UE techniques in a unified manner. Using this benchmark, we have conducted experiments with most methods implemented in LM-Polygraph. Below, we provide experimental details.

\paragraph{Datasets.}
  We experiment with three text generation tasks: machine translation (MT), text summarization (TS), and question answering (QA). For each task, we use two widely-used datasets: WMT-14 German to English and WMT-14 French to English~\cite{wmt14} for MT, XSum~\cite{xsum} and AESLC~\cite{aeslc} for TS, and CoQA~\cite{coqa} and bAbI QA~\cite{babiqa} for QA. Dataset statistics are presented in Appendix \ref{sec:datasets_stats}.

\paragraph{Models.}
  We conducted experiments with the Vicuna-v1.5-7B~\cite{vicuna} and Llama-v2-7B~\cite{touvron2023llama} models. The generation hyperparameters are provided in Appendix~\ref{sec:hyperopts}.

\paragraph{Metrics.}
  We focus on the task of selective generation~\cite{ren2023outofdistribution} where we ``rejecting'' generated sequences due to low quality based on uncertainty scores. Rejecting means that we do not use the model output, and the corresponding queries are processed differently: they could be further reprocessed manually or sent to a more advanced LLM.
  Following previous work on UE in text generation~\cite{malinin2020uncertainty,vazhentsev-etal-2022-uncertainty}, we compare the methods using the Prediction Rejection Ratio (PRR) metric~\cite{malinin-etal-2017-incorporating}.

  Consider a test dataset $\DC = \{(\xv_i, \yv_i)\}$. Let $f(\xv_i)$ be the output generated by an LLM and $U(\xv_i)$ be the uncertainty score of a prediction. 
  The prediction rejection (PR) curve indicates the dependence of the average quality $Q(f(\xv_i), \yv_i)$ of the covered instances from the uncertainty rate $a$ used for rejection, in ascending order. We use ROUGE-L and BERTScore~\cite{bertscore} as text quality metrics $Q(f(\xv_i), \yv_i)$. 
  Finally, PRR computes the ratio of the area $AUCPR_{unc}$ between the PR curve for the uncertainty estimates and random estimates and the area $AUCPR_{oracle}$ between the oracle and random estimates:
  \begin{equation}
    \label{eq:prr}
    PRR = \frac{AUCPR_{unc}}{AUCPR_{oracle}}
  \end{equation}
Higher PRR values indicate better quality of selective generation.
  

\section{Experimental Results}
\label{sec:results}
  Tables \ref{tab:vicuna_results} and \ref{tab:llama_2_results} present the results for Vicuna-v1.5-7b and LLaMA-v2-7b correspondingly.




  For both models, the better performance is usually demonstrated by the white-box methods based on information-theoretic concepts (first 8 rows of the table). These methods are in general also easy to implement and computationally lightweight, with the notable exceptions of Semantic Entropy, Monte Carlo Sequence Entropy, and Monte Carlo Normalized Sequence Entropy, which require sampling from the model several times to obtain uncertainty scores.

  When working with LLMs as web services, usually there is no access to full posterior distributions over tokens, therefore, only black-box methods could be used. Among this group of approaches, the best average performance is achieved by Eccentricity for Vicuna. For LLaMA, there is no clear advantage for any of the methods considered.

  Overall, we see 
  that absolute values for all evaluated methods, models, and datasets are far away from perfect. Low performance of current methods is especially evident on more complicated tasks such as XSum and WMT14. Our experimental results demonstrate that the task of selective generation is not close to be solved. This once again underlines the importance of further research and development of efficient uncertainty estimation techniques for generative language models.

\section{Conclusion}

As the community strives to advance the potential of LLMs, it is critical to be mindful about dangers of their uncontrolled usage. In this work, we propose a tool for making the application of LLMs safer. Enriching model predictions with uncertainty scores helps users and developers to be informed about these risks, encouraging healthy skepticism towards certain outputs generated by these models. 

We plan to further expand our framework with implementations of new UE methods that emerge in the future.
We hope that our work will foster the development of techniques to detect and mitigate LLM hallucinations, which we believe is a key to unlocking the safe, responsible, and effective use of LLMs in real-world applications.

\section*{Limitations}
  We have tried to be as comprehensive as possible with our collection of UE methods. However, we omit several techniques that have not demonstrated strong performance in previous work, do not have a strong theoretical motivation, or are similar to other implemented techniques. 

  We note that comprehensive evaluation of UE methods is an open research question. LM-Polygraph makes the first steps to systematize, and provide interfaces and tools for testing UE techniques in a unified manner. However, we believe that the number of tasks and datasets should be extended in the future. 

  When running the demo, we cannot provide an access to the biggest and the most powerful public LLMs, because running them is prohibitively expensive. Nevertheless, a user can access models such as ChatGPT by providing an API access key.

  LM-Polygraph supports common application program interfaces used by modern LLMs. However, it is possible that certain modifications will be required to support future releases of LLMs. 

  At the moment of writing, LM-polygraph provides valid uncertainty estimates only for model outputs in English language. This is due to the fact that most generation quality metrics implemented are based off English-specific implementations and non-multilingual models. We plan to alleviate this limitation by allowing the user to easily employ custom quality metrics and scoring models.

\section*{Ethics Statement}
  We conducted all experiments on publicly-available datasets that have been leveraged in various previous work on uncertainty estimation of LLMs. 

  While training data for most LLMs, such as BLOOMz, was selected to contain little or no abusive text content, such models can still potentially output harmful textual content. Techniques investigated in our work estimate certainty of an LM output to ``censor'' its output, and model debiasing is an orthogonal direction to our line of work. These additional methods can and perhaps should be combined in real production LLM deployments. We hope that our framework contributes to safer and more reliable usage of language models.

\section*{Acknowledgements}
We thank the anonymous reviewers for their insightful feedback towards improving this paper.

\bibliography{anthology,custom}
\bibliographystyle{acl_natbib}

\newpage
\clearpage

\appendix

\section{Methods Description}
\label{sec:appendix_methods}
  Here, we summarize UE methods implemented in LM-Polygraph; see also Table~\ref{tab:ue_methods}.

\subsection{White-box Methods}
\subsubsection{Information-based methods}
  \textit{Maximum sequence probability} score simply leverages the probability of the most likely sequence generation:
  $\mathrm{MSP}(\yv \mid \xv, \thetav) = 1 - P(\yv \mid \xv, \thetav)$.

  \textit{Length-normalized log probability} computes the average negative log probability of generated tokens. If the score is exponentiated it corresponds to \textit{perplexity}. 
  The resulting quantity is computed as
  \begin{EQA}[c]
    \mathrm{P}(\yv, \xv; \thetav) = \exp\Bigl\{-\frac{1}{L} \log P(\yv \mid \xv, \thetav)\Bigr\},
  \label{eq:nsp}
  \end{EQA}
  while it is convenient also to denote length-normalized sequence probability by $\bar{P}(\yv \mid \xv, \thetav) = \exp\Bigl\{\frac{1}{L} \log P(\yv \mid \xv, \thetav)\Bigr\}$.

  We also provide the \textit{mean token entropy}, where we simply average entropy of each individual token in the generated sequence:
  \begin{EQA}[c]
    \HC_T(\yv, \xv; \thetav) = \frac{1}{L} \sum\nolimits_{l = 1}^L \HC(y_l \mid \yv_{<l}, \xv, \thetav),
  \label{eq:entropy}
  \end{EQA}
  where $\HC(y_l \mid \yv_{<l}, \xv, \thetav)$ is an entropy of the token distribution $P(y_l \mid \yv_{<l}, \xv, \thetav)$.

  The other possibility to compute entropy-based uncertainty measure is to compute it on the level of whole sequences via $\EE \bigl[-\log P(\yv \mid \xv, \thetav)\bigr]$, where expectation is taken over the sequences $\yv$ randomly generated from the distribution $P(\yv \mid \xv, \thetav)$. In practice, one needs to use Monte-Carlo integration, i.e. generate several sequences $\yv^{(k)}, ~ k = 1, \dots, K$ via randoms sampling and compute the resulting \textit{Monte Carlo Sequence Entropy}:
  \begin{EQA}[c]
    \HC_S(\xv; \thetav) = -\frac{1}{K} \sum\nolimits_{k = 1}^K \log P(\yv^{(k)} \mid \xv, \thetav).
  \label{eq:seq_entropy}
  \end{EQA}
  The same procedure can be done by substituting $P(\yv^{(k)} \mid \xv, \thetav)$ with its length-normalized version $\bar{P}(\yv^{(k)} \mid \xv, \thetav)$ leading to a more reliable uncertainty measure in some applications.

  Another entropy-based uncertainty measure is \textit{Semantic Entropy} proposed by~\citet{kuhn2023semantic}. 
  The method aims to deal with the generated sequences that have similar meaning while having different probabilities according to the model, which can significantly affect the resulting entropy value~\eqref{eq:seq_entropy}. The idea is to cluster generated sequences $\yv^{(k)}, ~ k = 1, \dots, K$ into several semantically homogeneous clusters $\CC_m, ~ m = 1, \dots, M$ with $M \le K$ with bi-directional entailment algorithm and average the sequence probabilities within the clusters. The resulting estimate of entropy is given by the following formula:
  \begin{EQA}[c]
    \mathrm{SE}(\xv; \thetav) = -\sum\nolimits_{m = 1}^M \hat{P}_m(\xv; \thetav) \log \hat{P}_m(\xv; \thetav),
  \end{EQA}
  where $\hat{P}_m(\xv; \thetav) = \frac{1}{|\CC_m|} \sum_{\yv \in \CC_m} P(\yv \mid \xv, \thetav)$.

  Finally, one can consider negative mean \textit{Pointwise Mutual Information} (PMI; \citet{takayama-arase-2019-relevant}) which is given by
  \begin{EQA}[c]
    \mathrm{PMI}(\yv, \xv; \thetav) = \frac{1}{L} \sum\nolimits_{l = 1}^L \log \frac{P(y_l \mid \yv_{<l}, \thetav)}{P(y_l \mid \yv_{<l}, \xv, \thetav)}.
  \end{EQA}
  This method was extended in~\cite{van-der-poel-etal-2022-mutual} by considering only those marginal probabilities for which the entropy of the conditional distribution is above certain threshold: $\HC(y_l \mid \yv_{<l}, \xv, \thetav) \ge \tau$. It leads to the negative mean \textit{Conditional Pointwise Mutual Information (CPMI)} measure that is given by:
  \begin{EQA}
    && \!\!\!\! \mathrm{CPMI}(\yv, \xv; \thetav) = -\frac{1}{L} \sum\nolimits_{l = 1}^L \log P(y_l \mid \yv_{<l}, \xv, \thetav) \\
    && \quad + \frac{\lambda}{L} \sum\nolimits_{l\colon \HC(y_l \mid \yv_{<l}, \xv, \thetav) \ge \tau} \log P(y_l \mid \yv_{<l}, \thetav),
  \end{EQA}
  where $\lambda > 0$ is another tunable parameter.

\subsubsection{Ensemble-based methods}
  For the ensembling on a sequence level, we consider two uncertainty measures: total uncertainty measured via average sequence probability $\bar{P}(\yv \mid \xv) = \frac{1}{M} \sum_{i = 1}^M \bar{P}(\yv \mid \xv, \thetav_i)$:
  \begin{equation}
    \mathrm{MSP}_S(\yv, \xv) = 1 - \bar{P}(\yv \mid \xv)
  \label{eq:tu_seq} 
  \end{equation}
  and
  \begin{equation}
    \MC_S(\yv, \xv) = \frac{1}{M} \sum\nolimits_{i = 1}^M \log \frac{P(\yv \mid \xv)}{P(\yv \mid \xv, \thetav_i)},
  \label{eq:rmi_seq} 
  \end{equation}
  which is known as \textit{reverse mutual information (RMI)}. 



  Next we discus token level uncertainty measures and start with a total uncertainty estimate via entropy:
  \begin{equation}
    \HC_T(\yv, \xv) = \sum\nolimits_{l = 1}^{L} \HC(y_l \mid \yv_{<l}, \xv),
  \label{eq:tu_toc} 
  \end{equation}
  where $\HC(y_l \mid \yv_{<l}, \xv)$ is an entropy of the token distribution $P(y_l \mid \yv_{<l}, \xv) = \frac{1}{M} \sum_{i = 1}^M P(y_l \mid \yv_{<l}, \xv; \thetav_i)$. 

  Additionally, for the ensemble one can compute the variety of other token level uncertainty measures including average entropy of ensemble members (also known as~\textit{Data Uncertainty}):
  \begin{EQA}[c]
    \DC(y_l \mid \yv_{<l}, \xv) = \frac{1}{M} \sum\nolimits_{i = 1}^M \HC(y_l \mid \yv_{<l}, \xv, \thetav_i),
  \end{EQA}
  \textit{Mutual Information (MI)}:
  \begin{EQA}[c]
  \label{eq:mi_toc}
    \IC(y_l \mid \yv_{<l}, \xv) = \HC(y_l \mid \yv_{<l}, \xv) - \DC(y_l \mid \yv_{<l}, \xv)
  \end{EQA}
  and \textit{Expected Pairwise KL Divergence (EPKL)}:
  \begin{EQA}
  \label{eq:epkl_toc}
    && \KC(y_l \mid \yv_{<l}, \xv) = \binom{M}{2}^{-1} \cdot \\ 
    && \cdot \sum_{i \neq j} 
    \mathcal{KL}\bigl(P(y_l \mid \yv_{<l}, \xv, \thetav_i) ~\|~ P(y_l \mid \yv_{<l}, \xv, \thetav_j)\bigr),
  \end{EQA}
  where $\mathcal{KL}(P ~\|~ Q)$ refers to a KL-divergence between distributions $P$ and $Q$.

  Finally, \textit{Reverse Mutual Information (RMI)} also can be computed on the token level via a simple equation
  \begin{EQA}[c]
  \label{eq:rmi_toc}
    \MC(y_l \mid \yv_{<l}, \xv) = \KC(y_l \mid \yv_{<l}, \xv) - \IC(y_l \mid \yv_{<l}, \xv).
  \end{EQA}

  The resulting token-level uncertainties computed via Data Uncertainty, MI, EPKL and RMI can be plugged-in in equation~\eqref{eq:tu_toc} on the place of entropy leading to corresponding sequence level uncertainty estimates. 


\subsubsection{Density-based Methods}
  Let $h(\xv)$ be a hidden representation of an instance $\xv$. The \textit{Mahalanobis Distance} (MD; \citet{lee2018simple}) method fits a Gaussian centered at the training data centroid $\mu$ with an empirical covariance matrix $\Sigma$. The uncertainty score is the Mahalanobis distance between $h(\xv)$ and $\mu$:
  \begin{EQA}[c]
    \mathrm{MD}(\xv) = \bigl(h(\xv) - \mu\bigr)^{T} \Sigma^{-1} \bigl(h(\xv) - \mu\bigr).
  \end{EQA}
  We suggest using the last hidden state of the encoder averaged over non-padding tokens or the last hidden state of the decoder averaged over all generated tokens as $h(\xv)$.
  
  The {Robust Density Estimation} (RDE; \citet{yoo-etal-2022-detection}) method improves over MD by reducing the dimensionality of $h(\xv)$ via PCA decomposition. Additionally, computing of the covariance matrix $\Sigma$ for each individual class is done by using the Minimum Covariance Determinant estimation~\cite{Rousseeuw84leastmedian}. The uncertainty score is computed as the Mahalanobis distance between but in the space of reduced dimensionality.

  \citet{ren2023outofdistribution} showed that it might be useful to adjust the Mahalanobis distance score by subtracting from it the other Mahalanobis distance $\mathrm{MD}_0(\xv)$ computed for some large general purpose dataset covering many domains like C4~\cite{raffel2020exploring}. The resulting resulting \textit{Relative Mahalanobis Distance} score is
  \begin{EQA}[c]
    \mathrm{RMD}(\xv) = \mathrm{MD}(\xv) - \mathrm{MD}_0(\xv).
  \end{EQA}

\subsection{Black-box Methods}
  In this work, we follow~\citet{lin2023generating} and consider two approaches to compute the similarity for the generated responses. The first one is \textit{Jaccard similarity}:
  \begin{EQA}[c]
    s(\yv, \yv') = \frac{|\yv \cap \yv'|}{|\yv \cup \yv'|},
  \end{EQA}
  where the sequences $\yv$ and $\yv'$ are considered just as sets of words.

  The other similarity measure considered is Natural Language Index (NLI) which employs a classification model to identify whether two responses are similar. We follow~\citet{kuhn2023semantic} and use the DeBERTa-large model~\cite{he2020deberta} that, for each pair of input sequences, provides two probabilities: $\hat{p}_{\mathrm{entail}}(\yv, \yv')$ that measures the degree of entailment between the sequences and $\hat{p}_{\mathrm{contra}}(\yv, \yv')$ that measures the contradiction between them. Then one can use $s_{\mathrm{entail}}(\yv, \yv') = \hat{p}_{\mathrm{entail}}(\yv, \yv')$ or $s_{\mathrm{contra}}(\yv, \yv') = 1 - \hat{p}_{\mathrm{contra}}(\yv, \yv')$ as a measure of similarity between sequences $\yv$ and $\yv'$.


  \textit{Number of Semantic Sets} illustrates whether answers are semantically equivalent. We adopt an iterative approach by sequentially examining responses from the first to the last while making pairwise comparisons between them (each pair has indexes $j_1$ and $j_2$, $j_2 > j_1$). The number of semantic sets initially equals the total number of generated answers $K$. If the condition $\hat{p}_{\mathrm{entail}}(\yv_{j_1},\yv_{j_2}) > \hat{p}_{\mathrm{contra}}(\yv_{j_1},\yv_{j_2}) $ and $\hat{p}_{\mathrm{entail}}(\yv_{j_2},\yv_{j_1}) > \hat{p}_{\mathrm{contra}}(\yv_{j_2}, \yv_{j_1}) $ is fulfilled we put this two sentences into one cluster. The computation is done for all the pairs of answers, and then the resulting number of distinct sets $U_{NumSemSets}$ is reported.
  It is worth noting that a higher number of semantic sets corresponds to an increased level of uncertainty, as it suggests a higher number of diverse semantic interpretations for the answer.
  

  Nonetheless, it is essential to acknowledge a limitation of this measure: it can only take integer values. Additionally, it cannot be assumed that the semantic equivalence derived from the NLI model is always transitive. Consequently, the authors of~\cite{lin2023generating} suggest the consideration of a continuous counterpart of this metric. They propose the \textit{Sum of Eigenvalues of the Graph Laplacian} as a potential alternative approach.
  
  Let's consider a similarity matrix $S_{j_1 j_2} = \bigl(s(\yv_{j_1}, \yv_{j_2}) + s(\yv_{j_2}, \yv_{j_1})\bigr) / 2$. Averaging is done to obtain better consistency. 
  Normalized Graph Laplacian of the obtained similarity Matrix $S$ has the following formula $L = I - D^{-\frac{1}{2}} S  D^{-\frac{1}{2}}$, where $D$ is a diagonal matrix and $D_{ii} = \sum_{j = 1}^K{S_{ij}}$. 
  %
  Consequently, the following formula is derived: $U_{EigV} = \sum_{k = 1}^{K}{\max(0, 1 - \lambda_{k})}$. This value is a continuous analogue of $U_{NumSemSets}$. In extreme case if adjacency matrix $S$ is binary these two measures will coincide.
  

  Of course, from a theoretical and practical point of view, $U_{EigV}$ is a much more flexible approach compared to $U_{NumSemSets}$. Still, they have a common disadvantage: they can not provide uncertainty for each answer. However, authors of~\cite{lin2023generating} demonstrate that we can take it from \textit{Degree Matrix} $D$ computed above. The idea is that the total uncertainty of the answers might be measured as a corrected trace of the diagonal matrix $D$ because elements on the diagonal of matrix $D$ are sums of similarities between the given answer and other answers. Thus, it is an average pairwise distance between all answers, and a larger value will indicate larger uncertainty because of the larger distance between answers. The resulting uncertainty measure becomes $U_{Deg} = 1 - trace(D) / K^{2}$. 
 

  A drawback of previously considered methods is the limited knowledge of the actual embedding space for the different answers since we only have measures of their similarities. Nevertheless, we can overcome this limitation by taking advantage of the inferential capabilities of the graph Laplacian, which makes it easier to obtain the coordinates of the answers. Let us introduce $\uv_1, \dots, \uv_{k} \in R^{K}$ as the eigenvectors of $L$ that correspond to $k$ smallest eigenvalues. We can efficiently construct an informative embedding $\vv_{j} = [\uv_{1,j}, \dots, \uv_{k,j}]$ for an answer $\yv_{j}$. Authors of~\cite{lin2023generating} demonstrate that this approach allows the usage of the average distance from the center as an uncertainty metric and to consider the distance of each response from the center as a measure of (negative) confidence. In mathematical terms, the estimates for \textit{Eccentricity} can be defined as follows: $U_{Ecc} = \bigl\|[\tilde{\vv}_{1}^{T}, \dots, \tilde{\vv}_{K}^{T}]\bigr\|_{2}$, where $\tilde{\vv}_{j} = \vv_{j} - \frac{1}{K}\sum_{\ell = 1}^{K}{\vv_{\ell}}$. 
 
  
  Last but not least, \textit{Lexical Similarity} is a measure proposed by~\cite{fomicheva-etal-2020-unsupervised} that computes how similar two words or phrases are in terms of their meaning. Since the original article is dedicated to machine translation, this measure calculates the average similarity score between all pairs of translation hypotheses in a set, using a similarity measure based on the overlap of their lexical items. Different metrics can be used, such as ROUGE-1, ROUGE-2, ROUGE-L, and BLEU. For our task, this measure iterates over all responses and calculates the average score with other answers. 


\clearpage
\FloatBarrier
\onecolumn
\section{Generation Hyperparameters}
\label{sec:hyperopts}

\begin{table*}[h!] 
\centering\resizebox{\textwidth}{!}{
\begin{tabular}{l|c|c|c|c|c|c|c|c}
\toprule
\textbf{Dataset} & \textbf{Task} & \textbf{Max Input Length} & \textbf{Generation Length} & \textbf{Temperature} & \textbf{Top-p} & \textbf{Do Sample} & \textbf{Beams} & \textbf{Repetition Penalty} \\
\midrule\midrule
 AESLC & \multirow{2}{*}{ATS} & \multirow{6}{*}{2048} & 31 & \multirow{6}{*}{1.0} & \multirow{6}{*}{1.0} & \multirow{6}{*}{False} & \multirow{6}{*}{1} & \multirow{6}{*}{1} \\
 XSUM &  &  & 56 &  &  &  &  &  \\
 CoQA & \multirow{2}{*}{QA} &  & 20 &  &  &  &  & \\
 bAbiQA &  &  & 3 &  &  &  &  & \\
 WMT14 De-En & \multirow{2}{*}{NMT} &  & 107&  &  &  &  & \\
 WMT14 Fr-En &  &  & 107 &  &  &  &  & \\
\bottomrule
\end{tabular}}
\caption{\label{tab:hyperparameters} Text generation hyperparameters for both LLMs Vicuna-v1.5-7b and Llama-2-7b used in the experiments.}
\end{table*}

Table~\ref{tab:hyperparameters} presents the hyperparameters used for experiments with LLMs Vicuna-v1.5-7b and LLaMA-2-7b-hf on various datasets and tasks. 
Maximum length of generated sequence was set for each dataset as the 99th percentile of target sequence length on the respecitve train set.


\section{Text Generation Quality Metrics}
\label{sec:quality_metrics}

\begin{table*}[!ht] \resizebox{\textwidth}{!}{\begin{tabular}{cc|cc|cc|cc|cc|cc}
\toprule
\multicolumn{2}{c|}{\textbf{AESLC}} & \multicolumn{2}{c|}{\textbf{XSUM}} & \multicolumn{2}{c|}{\textbf{CoQA}} & \multicolumn{2}{c|}{\textbf{bAbiQA}} & \multicolumn{2}{c|}{\textbf{WMT14 De-En}} & \multicolumn{2}{c}{\textbf{WMT14 Fr-En}} \\ \midrule
    \textbf{Rouge-L} & \textbf{BERTScore}& \textbf{Rouge-L} & \textbf{BERTScore}& \textbf{Rouge-L} & \textbf{BERTScore}& \textbf{Rouge-L} & \textbf{BERTScore}&  \textbf{Rouge-L} & \textbf{BERTScore}& \textbf{Rouge-L} & \textbf{BERTScore} \\\midrule
\midrule
   0.24 &      0.83 &    0.18 &      0.86 &    0.29 &      0.85 &    0.68 &       1.0 &       0.59 &      0.95 &       0.64 &      0.95 \\
\bottomrule
\end{tabular}
}\caption{\label{tab:vicuna_quality} Rouge-L$\uparrow$ and BERTScore$\uparrow$ for Vicuna v1.5 model for various tasks.}\end{table*}
\begin{table*}[!ht] \resizebox{\textwidth}{!}{\begin{tabular}{cc|cc|cc|cc|cc|cc}
\toprule
\multicolumn{2}{c|}{\textbf{AESLC}} & \multicolumn{2}{c|}{\textbf{XSUM}} & \multicolumn{2}{c|}{\textbf{CoQA}} & \multicolumn{2}{c|}{\textbf{bAbiQA}} & \multicolumn{2}{c|}{\textbf{WMT14 De-En}} & \multicolumn{2}{c}{\textbf{WMT14 Fr-En}} \\ \midrule
    \textbf{Rouge-L} & \textbf{BERTScore}& \textbf{Rouge-L} & \textbf{BERTScore}& \textbf{Rouge-L} & \textbf{BERTScore}& \textbf{Rouge-L} & \textbf{BERTScore}&  \textbf{Rouge-L} & \textbf{BERTScore}& \textbf{Rouge-L} & \textbf{BERTScore} \\\midrule
\midrule
   0.23 &      0.84 &    0.19 &      0.86 &    0.51 &      0.91 &    0.36 &      0.98 &       0.54 &      0.93 &       0.56 &      0.92 \\
\bottomrule
\end{tabular}
}\caption{\label{tab:llama_quality} Rouge-L$\uparrow$ and BERTScore$\uparrow$ for the Llama v2 model for various tasks.}\end{table*}

\section{Dataset Statistics}
\label{sec:datasets_stats}

\begin{table*}[ht]
\centering
\resizebox{0.8\textwidth}{!}{\begin{tabular}{cccccc}
\toprule
\textbf{Dataset} & \textbf{Num. instances} & \textbf{Av. document len.} & \textbf{Av. target len.} & \textbf{Language} &  \\
\hline
\hline
\multicolumn{5}{c}{\textbf{NMT}} \\
WMT'14 & 4.51M / 3000 / \textbf{3003} & 19.8 / 18.3 & 23.0 / 21.3 & German-to-English \\
WMT'14 & 40.8M / 3000 / \textbf{3003} & 33.5 / 32.1 & 29.2 / 27.0 & French-to-English \\\hline
\multicolumn{5}{c}{\textbf{ATS}} \\
XSum & 204045 / 11332 / \textbf{11334} & 454.6 & 26.1 & English \\ 
AESLC & 14436 / 1960 / \textbf{1906} & 165.5 & 6.7 & English \\ \hline
\multicolumn{5}{c}{\textbf{QA}} \\
CoQA & 7199 / 500 / - & 271.4 & 2.7 & English \\ 
bAbiQA & 2000 / - / \textbf{200} & 31.1 & 1.0 & English \\ \bottomrule

\end{tabular}}
\caption{Quantitative information regarding the datasets from experiments. It includes the count of instances available for the training, validation, and \textbf{test} sets, as well as the mean lengths of both texts and targets (answers / translations / summaries) measured in terms of tokens. In addition, the languages of the source and target texts are also specified.}
\label{tab:datasets_stats}
\end{table*}

Table~\ref{tab:datasets_stats} illustrates the statistics of the datasets that were used in the experiments. Experiments were conducted using all examples from the test sets of these datasets, while training density-based methods were performed on a random subset of 1000 elements from the train set.

\begin{figure}[!ht]
\begin{lstlisting}[language=bash]
HYDRA_CONFIG=/path/to/cloned/repo/examples/configs/polygraph_eval_coqa.yaml polygraph_eval model=lmsys/vicuna-7b-v1.5

\end{lstlisting}
\vspace{-0.3cm}
\caption{Script that reproduces benchmark results for CoQA dataset with Vicuna-v1.5-7b model.}\label{code:script}
\end{figure}

To evaluate the performance of considered uncertainty estimation methods, we provide code to retrieve benchmark results. Figure~\ref{code:script} shows an example of starting an experiment with the Vicuna-v1.5-7b model on the Questions Answering task (CoQA dataset).

\begin{figure}[!ht]
\begin{lstlisting}[language=yaml]
hydra:
  run:
    dir: ${cache_path}/${task}/${model}/${dataset}/${now:%Y-%m-%d}/${now:%H-%M-%S}

cache_path: ./workdir/output
save_path: '${hydra:run.dir}'

device: cpu

task: qa

dataset: coqa
text_column: questions
label_column: answers
prompt: "Answer a question given a story. Output only the answer.\nStory:\n{story}\n\nQuestion:\n{question}\n\nAnswer:\n"
train_split: train
eval_split: validation
max_new_tokens: 20
load_from_disk: false

train_dataset: null
train_test_split: false
test_split_size: 1

background_train_dataset: allenai/c4
background_train_dataset_text_column: text
background_train_dataset_label_column: url
background_train_dataset_data_files: en/c4-train.00000-of-01024.json.gz
background_load_from_disk: false

subsample_background_train_dataset: 1000
subsample_train_dataset: 1000
subsample_eval_dataset: -1

model: lmsys/vicuna-7b-v1.5
use_auth_token:

use_density_based_ue: true
use_seq_ue: true
use_tok_ue: false

ignore_exceptions: false

batch_size: 1
deberta_batch_size: 10

seed:
    - 1
    
\end{lstlisting}
\caption{Config Example for Question Answering on CoQA dataset.}
\label{code:config}
\end{figure}

Figure~\ref{code:config} shows an example of a config file used for experiment related to CoQA dataset with Vicuna-v1.5-7b model. It contains information about import and parameters. For other datasets and models the config structure is the same.

\section{Normalization of Uncertainty Estimates in Demo App}
\label{sec:calibraion_normalization}
To make uncertainty estimation more intuitive for the end user, directly interacting with the LLM, we perform normalization of various uncertainty estimates. After normalization the output $\mathrm{UE}(\xv)$ of any uncertainty estimation approach becomes a confidence score $\mathrm{C}(\xv) \in [0, 1] \subset \mathbb{R}$. 

We experimented with several ways of achieving this normalization, including quantile-based approach and simple linear normalization on maximum value obtained from validation dataset. Eventually we performed normalization as a calibration procedure, where normalized confidence score represents expected value of generation quality metric of choice (i.e. RougeL) for a given uncertainty estimate. This expectation is estimated by computing sample averages of quality metric over bins of uncertainty estimates, calculated for some validation dataset. For RougeL metric, the confidence estimate $\mathrm{C}(\xv_{input})$ thus becomes:

  \begin{EQA}[c]
    \mathrm{C}(\xv_{input}) = \frac{1}{|\mathcal{B}|}\sum_{\xv_i, \yv_i \in \mathcal{B}} rougeL(\hat{\yv}_i, \yv_i),
  \label{eq:calibration}
  \end{EQA}

where $\hat{\yv}_i$ is model output for input $\xv_i$, and 

  \begin{EQA}[c]
    \mathcal{B} = \{(\xv, \yv) \in \mathcal{D}_{calib} \mid \mathrm{UE}(\xv) \in [\mathrm{UE}_{\min}, \mathrm{UE}_{\max}) \}
  \label{eq:bins}
  \end{EQA}
is the bin to which uncertainty estimate of the input belongs. The bounds of this bin are selected from the predetermined set of bin boundaries to be the neighboring pair for which condition

  \begin{EQA}[c]
    \mathrm{UE}_{\min} \le \mathrm{UE}(\xv_{input}) < \mathrm{UE}_{\max}
  \label{eq:bin_bounds}
  \end{EQA}
is satisfied.

This dataset $\mathcal{D}_{calib}$ is constructed to be representative of different modes of operation of a given model. For this purpose it is constructed as a mixture of several different datasets for different tasks, with different values of relevant statistics, such as input sequence length, typical generated output length etc. 

It is obvious that quality of this normalized confidence score depends heavily on the size and diversity of the calibration dataset. In general we consider the problem of translating opaque uncertainty estimates into intuitive absolute confidence scores, that correctly represent likelihood of the generated output being correct and relevant, as an important and complicated task. We leave solving this problem in a more efficient and universal way to the future work.

\end{document}